\documentclass{article}
\pdfoutput = 1
\PassOptionsToPackage{numbers, compress}{natbib}

\usepackage[final]{nips_2018}

\usepackage[utf8]{inputenc} 
\usepackage[T1]{fontenc}    
\usepackage[breaklinks=true,colorlinks,bookmarks=false]{hyperref}       
\usepackage{url}            
\usepackage{booktabs}       
\usepackage{amsfonts}       
\usepackage{nicefrac}       
\usepackage{microtype}      
\usepackage[usenames, dvipsnames]{color}
\usepackage{graphicx}
\usepackage{url}
\title{SNIPER: Efficient Multi-Scale Training}

\author{
  Bharat Singh $^*$  \hspace{1.5cm}  Mahyar Najibi \thanks{Equal Contribution} \hspace{1.5cm}  Larry S. Davis \\
  University of Maryland, College Park\\
  \{\texttt{bharat,najibi,lsd}\}\texttt{@cs.umd.edu} 
}

\begin{document}

\maketitle

\begin{abstract}
We present \textit{SNIPER}, an algorithm for performing efficient multi-scale training in instance level visual recognition tasks. Instead of processing every pixel in an image pyramid, SNIPER processes context regions around ground-truth instances (referred to as \textit{chips}) at the appropriate scale. For background sampling, these context-regions are generated using proposals extracted from a region proposal network trained with a short learning schedule. Hence, the number of chips generated per image during training adaptively changes based on the scene complexity. SNIPER only processes 30\% more pixels compared to the commonly used single scale training at 800x1333 pixels on the COCO dataset. But, it also observes samples from extreme resolutions of the image pyramid, like 1400x2000 pixels. As SNIPER operates on resampled low resolution chips (512x512 pixels), it can have a batch size as large as $20$ on a single GPU even with a ResNet-101 backbone. Therefore it can benefit from batch-normalization during training without the need for synchronizing batch-normalization statistics across GPUs. SNIPER brings training of instance level recognition tasks like object detection closer to the protocol for image classification and suggests that the commonly accepted guideline that it is important to train on high resolution images for instance level visual recognition tasks might not be correct. Our implementation based on Faster-RCNN with a ResNet-101 backbone obtains an mAP of $47.6$\% on the COCO dataset for bounding box detection and can process $5$ images per second during inference with a single GPU. Code is available at \url{https://github.com/mahyarnajibi/SNIPER/}.
 
\end{abstract}

\section{Introduction}
Humans have a foveal visual system which attends to objects at a fixed distance and size. For example, when we focus on nearby objects, far away objects get blurred \cite{corbetta2002control}. Naturally, it is difficult for us to focus on objects of different scales simultaneously \cite{rensink2000dynamic}. We only process a small field of view at any given point of time and adaptively ignore the remaining visual content in the image. However, computer algorithms which are designed for instance level visual recognition tasks like object detection depart from this natural way of processing visual information. For obtaining a representation robust to scale, popular detection algorithms like Faster-RCNN/Mask-RCNN \cite{ren2015faster,he2017mask} are trained on a multi-scale image pyramid \cite{liu2018path,singh2017analysis}. Since every pixel is processed at each scale, this approach to processing visual information increases the training time significantly. For example, constructing a $3$ scale image pyramid (\textit{e.g.} scales=$1$x,$2$x,$3$x) requires processing $14$ times the number of pixels present in the original image. For this reason, it is impractical to use multi-scale training in many scenarios. 

Recently, it is shown that ignoring gradients of objects which are of extreme resolutions is beneficial while using multiple scales during training \cite{singh2017analysis}. For example, when constructing an image pyramid of $3$ scales, the gradients of large and small objects should be ignored at large and small resolutions respectively. If this is the case, an intuitive question which arises is, do we need to process the entire image at a $3$x resolution? Wouldn’t it suffice to sample a much smaller RoI (chip) around small objects at this resolution? On the other hand, if the image is already high resolution, and objects in it are also large in size, is there any benefit in upsampling that image? 

While ignoring significant portions of the image would save computation, a smaller chip would also lack context required for recognition. A significant portion of background would also be ignored at a higher resolution. So, there is a trade-off between computation, context and negative mining while accelerating multi-scale training. To this end, we present a novel training algorithm called \textit{Scale Normalization for Image Pyramids with Efficient Resampling (SNIPER)}, which adaptively samples chips from multiple scales of an image pyramid, conditioned on the image content. We sample positive chips conditioned on the ground-truth instances and negative chips based on proposals generated by a region proposal network. Under the same conditions (fixed batch normalization), we show that SNIPER performs as well as the multi-scale strategy proposed in SNIP \cite{singh2017analysis} while reducing the number of pixels processed by a factor of $3$ during training on the COCO dataset. Since SNIPER is trained on $512$x$512$ size chips, it can reap the benefits of a large batch size and training with batch-normalization on a single GPU node. In particular, we can use a batch size of $20$ per GPU (leading to a total batch size of 160), even with a ResNet-101 based Faster-RCNN detector. While being efficient, SNIPER obtains competitive performance on the COCO detection dataset even with simple detection architectures like Faster-RCNN. 

\section{Background}
Deep learning based object detection algorithms have primarily evolved from the R-CNN detector \cite{girshick2014rich}, which started with object proposals generated with an unsupervised algorithm \cite{uijlings2013selective}, resized these proposals to a canonical $224$x$224$ size image and classified them using a convolutional neural network \cite{lecun1998gradient}. This model is scale invariant, but the computational cost for training and inference for R-CNN scales linearly with the number of proposals. To alleviate this computational bottleneck, Fast-RCNN \cite{girshick2015fast} proposed to project region proposals to a high level convolutional feature map and use the pooled features as a semantic representation for region proposals. In this process, the computation is shared for the convolutional layers and only lightweight fully connected layers are applied on each proposal. However, convolution for objects of different sizes is performed at a single scale, which destroys the scale invariance properties of R-CNN. Hence, inference at multiple scales is performed and detections from multiple scales are combined by selecting features from a pair of adjacent scales closer to the resolution of the pre-trained network \cite{he2014spatial,girshick2015fast}. The Fast-RCNN model has since become the {\em de-facto} approach for classifying region proposals as it is fast and also captures more context in its features, which is lacking in RCNN.

It is worth noting that in multi-scale training, Fast-RCNN upsamples {\em and} downsamples every proposal (whether small or big) in the image. This is unlike R-CNN, where each proposal is resized to a canonical size of $224$x$224$ pixels. Large objects are not upsampled and small objects are not downsampled in R-CNN. In this regard, R-CNN more appropriately does not up/downsample every pixel in the image but only in those regions which are likely to contain objects to an appropriate resolution. However, R-CNN does not share the convolutional features for nearby proposals like Fast-RCNN, which makes it slow. To this end, we propose  SNIPER, which retains the benefits of both these approaches by generating scale specific context-regions (chips) that cover maximum proposals at a particular scale. SNIPER classifies all the proposals inside a chip like Fast-RCNN which enables us to perform efficient classification of multiple proposals within a chip. As SNIPER does not upsample the image where there are large objects and also does not process easy background regions, it is significantly faster compared to a Fast-RCNN detector trained on an image pyramid.


SNIP \cite{singh2017analysis} is also trained on almost all the pixels of the image pyramid (like Fast-RCNN), although gradients arising from objects of extreme resolutions are ignored. In particular, $2$ resolutions of the image pyramid ($480$ and $800$ pixels) always engage in training and multiple $1000$ pixel crops are sampled out of the $1400$ pixel resolution of the image in the finest scale. SNIPER takes this cropping procedure to an extreme level by sampling $512$ pixels crops from $3$ scales of an image pyramid. At extreme scales (like $3$x), SNIPER observes less than one tenth of the original content present in the image! Unfortunately, as SNIPER chips generated only using ground-truth instances are very small compared to the resolution of the original image, a significant portion of the background does not participate in training. This causes the false positive rate to increase. Therefore, it is important to generate chips for background regions as well. In SNIPER, this is achieved by randomly sampling a fixed number of chips (maximum of $2$ in this paper) from regions which are likely to cover false positives. To find such regions, we train a lightweight RPN network with a short schedule. The proposals of this network are used to generate chips for regions which are likely to contain false positives (this could potentially be replaced with unsupervised proposals like EdgeBoxes \cite{zitnick2014edge} as well). After adding negative chip sampling, the performance of SNIPER matches SNIP, but it is $3$ times faster! Since we are able to obtain similar performance by observing less than one tenth of the image, it implies that very large context during training is {\em not} important for training high-performance detectors but sampling regions containing hard negatives is. 

\section{SNIPER}
We describe the major components of SNIPER in this section. One is positive/negative chip mining and the other is label assignment after chips are generated. Finally, we will discuss the benefits of training with SNIPER.

\subsection{Chip Generation}
SNIPER generates chips $\mathcal{C}^i$ at multiple scales $\{s_1, s_2,.., s_i,.. s_n\}$ in the image. For each scale, the image is first re-sized to width ($W_i$) and height ($H_i$). On this canvas, $K \times K$ pixel chips are placed at equal intervals of $d$ pixels (we set $d$ to $32$ in this paper). This leads to a two-dimensional array of chips at each scale.

\subsection{Positive Chip Selection}

\begin{figure}
\centering{
\includegraphics[width=0.95\linewidth]{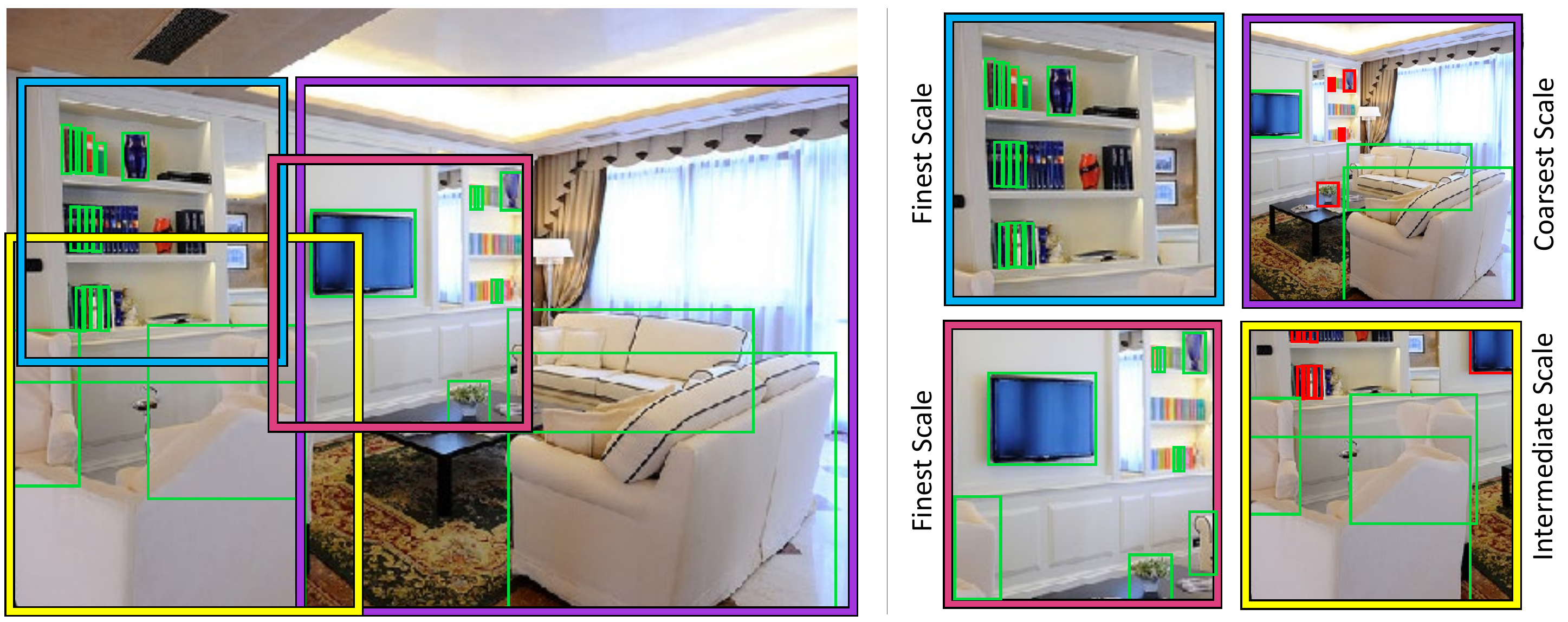}}
\caption{SNIPER Positive chip selection . SNIPER adaptively samples context regions (aka chips) based on the presence of objects inside the image. Left side: The image, ground-truth boxes (represented by green lines), and the chips in the original image scale (represented by the blue, yellow, pink, and purple rectangles). Right side: Down/up-sampling is performed considering the size of the objects. Covered objects are shown in green and invalid objects in the corresponding scale are shown as red rectangles. }
\label{fig:pos_chips}
\end{figure} 

For each scale, there is a desired area range $\mathcal{R}^i = [r_{min}^{i}, r_{max}^{i}]$, $i \in [1, n]$ which determines which ground-truth boxes/proposals participate in training for each scale $i$. The valid list of ground-truth bounding boxes which lie in $\mathcal{R}^i$ are referred to as $\mathcal{G}^i$. Then, chips are greedily selected so that maximum number of valid ground-truth boxes ($\mathcal{G}^i$) are covered. A ground-truth box is said to be covered if it is completely enclosed inside a chip. All the positive chips from a scale are combined per image and are referred to as $\mathcal{C}_{pos}^i$. For each ground-truth bounding box, there always exists a chip which covers it. Since consecutive $\mathcal{R}^i$ contain overlapping intervals, a ground-truth bounding box may be assigned to multiple chips at different scales. It is also possible that the same ground-truth bounding box may be in multiple chips from the same scale. Ground-truth instances which have a partial overlap (IoU > 0) with a chip are cropped. All the cropped ground-truth boxes (valid or invalid) are retained in the chip and are used in label assignment.

In this way, every ground-truth box is covered at the appropriate scale. Since the crop-size is much smaller than the resolution of the image (\textit{i.e.} more than $10$x smaller for high-resolution images), SNIPER does not process most of the background at high-resolutions. This leads to significant savings in computation and memory requirement while processing high-resolution images. We illustrate this with an example shown in Figure \ref{fig:pos_chips}. The left side of the figure shows the image with the ground-truth boxes represented by green bounding boxes. Other colored rectangles on the left side of the figure show the chips generated by SNIPER in the original image resolution which cover all objects. These chips are illustrated on the right side of the figure with the same border color. Green and red bounding boxes represent the valid and invalid ground-truth objects corresponding to the scale of the chip. As can be seen, in this example, SNIPER efficiently processes all ground-truth objects in an appropriate scale by forming $4$ low-resolution chips.

\subsection{Negative Chip Selection}
\label{sec:neg}
Although positive chips cover all the positive instances, a significant portion of the background is not covered by them. Incorrectly classifying background increases the false positive rate. In current object detection algorithms, when multi-scale training is performed, every pixel in the image is processed at all scales. Although training on all scales reduces the false positive rate, it also increases computation. We posit that a significant amount of the background is easy to classify and hence, we can avoid performing any computation in those regions. So, how do we eliminate regions which are easy to classify? A simple approach is to employ object proposals to identify regions where objects are likely to be present. After all, our classifier operates on region proposals and if there are no region proposals in a part of the image, it implies that it is very easy to classify as background. Hence, we can ignore those parts of the image during training.

\begin{figure}
\centering{
\includegraphics[width=\linewidth]{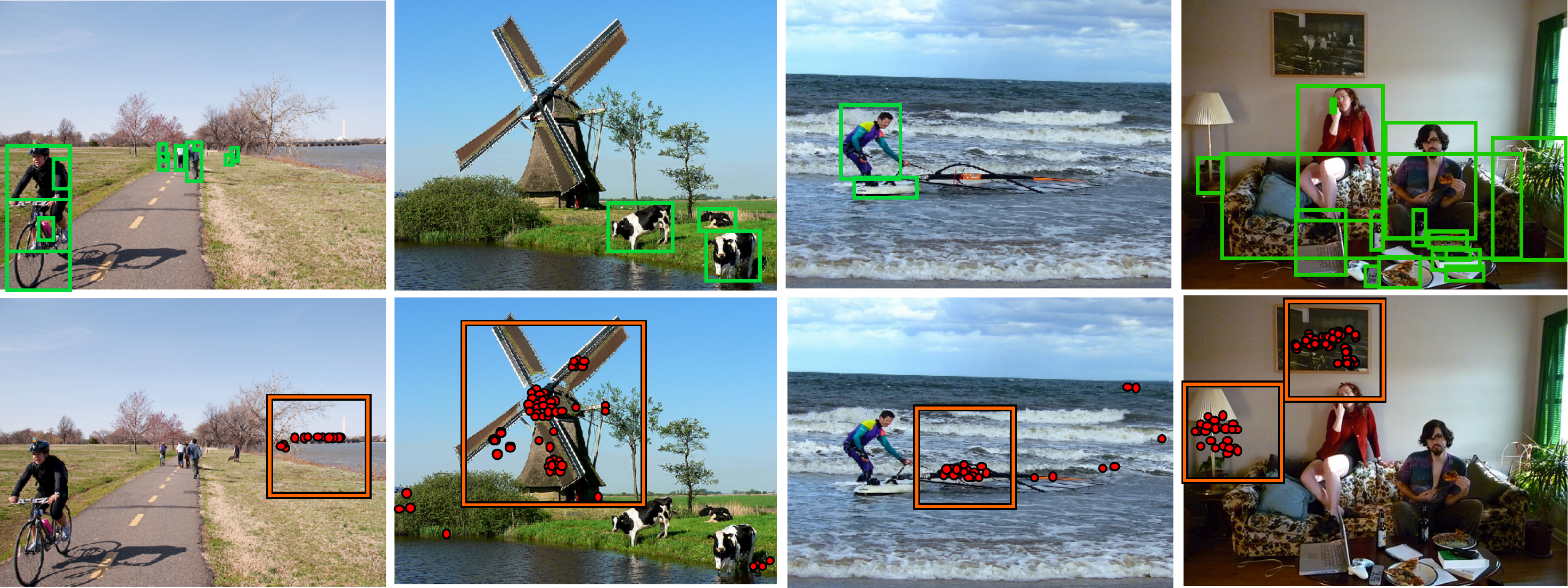}}
\caption{SNIPER negative chip selection. First row: the image and the ground-truth boxes. Bottom row: negative proposals not covered in positive chips (represented by red circles located at the center of each proposal for the clarity) and the generated negative chips based on the proposals (represented by orange rectangles).}
\label{fig:neg_chips}
\end{figure}

To this end, for negative chip mining, we first train RPN for a couple of epochs. No negative chips are used for training this network. The task of this network is to roughly guide us in selecting regions which are likely to contain false positives, so it is not necessary for it to be very accurate. This RPN is used to generate proposals over the entire training set.  We assume that if no proposals are generated in a major portion of the image by RPN, then it is unlikely to contain an object instance. For negative chip selection, for each scale $i$, we first remove all the proposals which have been covered in $\mathcal{C}_{pos}^i$. Then, for each scale $i$, we greedily select all the chips which cover at least $M$ proposals in $\mathcal{R}^i$. This generates a set of negative chips for each scale per image, $\mathcal{C}_{neg}^i$. During training, we randomly sample a fixed number of negative chips per epoch (per image) from this pool of negative chips which are generated from all scales, i.e. $\bigcup_{i=1}^{n} \mathcal{C}_{neg}^i$. Figure \ref{fig:neg_chips} shows examples of the generated negative chips by SNIPER. The first row shows the image and the ground-truth boxes. In the bottom row, we show the proposals not covered by $\mathcal{C}_{pos}^i$ and the corresponding negative chips generated (the orange boxes). However, for clarity, we represent each proposal by a red circle in its center. As illustrated, SNIPER only processes regions which likely contain false positives, leading to faster processing time.

\subsection{Label Assignment}
Our network is trained end to end on these chips like Faster-RCNN, i.e. it learns to generate proposals as well as classify them with a single network. While training, proposals generated by RPN are assigned labels and bounding box targets (for regression) based on {\em all} the ground-truth boxes which are present inside the chip. We do not filter ground-truth boxes based on $\mathcal{R}^i$. Instead, proposals which do not fall in $\mathcal{R}^i$ are ignored during training. So, a large ground-truth box which is cropped, could generate a valid proposal which is small. Like Fast-RCNN, we mark any proposal which has an overlap greater than 0.5 with a ground-truth box as positive and assign bounding-box targets for the proposal. Our network is trained end to end and we generate 300 proposals per chip. We do not apply any constraint that a fraction of these proposals should be re-sampled as positives \cite{ren2015faster}, as in Fast-RCNN. We did not use OHEM \cite{shrivastava2016training} for classification and use a simple softmax cross-entropy loss for classification. For assigning RPN labels, we use valid ground-truth boxes to assign labels and invalid ground-truth boxes to invalidate anchors, as done in SNIP \cite{singh2017analysis}. 

\subsection{Benefits}
For training, we randomly sample chips from the whole dataset for generating a batch. On average, we generate $\sim 5$ chips of size $512$x$512$ per image on the COCO dataset (including negative chips) when training on three scales ($512$/ms \footnote{max($width_{im}$,$height_{im}$)}, $1.667$, $3$). This is only $30$\% more than the number of pixels processed per image when single scale training is performed with an image resolution of $800$x$1333$. Since all our images are of the same size, data is much better packed leading to better GPU utilization which easily overcomes the extra $30$\% overhead. But more importantly, {\em we reap the benefits of multi-scale training on $3$ scales, large batch size and training with batch-normalization without any slowdown in performance on a single 8 GPU node!}.

It is commonly believed that high resolution images (e.g. $800$x$1333$) are necessary for instance level recognition tasks. Therefore, for instance level recognition tasks, it was not possible to train with batch-normalization statistics computed on a single GPU. Methods like synchronized batch-normalization \cite{liu2018path,zhao2017pyramid} or training on $128$ GPUs \cite{peng2017megdet} have been proposed to alleviate this problem. Synchronized batch-normalization slows down training significantly and training on $128$ GPUs is also impractical for most people. Therefore, group normalization \cite{GroupNorm2018} has been recently proposed so that instance level recognition tasks can benefit from another form of normalization in a low batch setting during training. With SNIPER, we show that the image resolution bottleneck can be alleviated for instance level recognition tasks. As long as we can cover negatives and use appropriate scale normalization methods, we can train with a large batch size of resampled low resolution chips, even on challenging datasets like COCO. Our results suggest that context beyond a certain field of view may not be beneficial during training. It is also possible that the effective receptive field of deep neural networks is not large enough to leverage far away pixels in the image, as suggested in \cite{luo2016understanding}.

In very large datasets like \textit{OpenImagesV4} \cite{openimages} containing $1.7$ million images, most objects are large and images provided are high resolution ($1024$x$768$), so it is less important to upsample images by $3\times$. In this case, with SNIPER, we generate $3.5$ million chips of size $512$x$512$ using scales of ($512$/ms, 1). Note that SNIPER also performs adaptive downsampling. Since the scales are smaller, chips would cover more background, due to which the impact of negative sampling is diminished. In this case (of positive chip selection), SNIPER processes only half the number of pixels compared to na\"ive multi-scale training on the above mentioned scales in OpenImagesV4. Due to this, we were able to train Faster-RCNN with a ResNet-101 backbone on $1.7$ million images in just $3$ days on a single $8$ GPU node!

\section{Experimental Details}
\label{sec:exp_details}
We evaluate SNIPER on the COCO dataset for object detection. COCO contains 123,000 images in the training and validation set and  20,288 images in the test-dev set. We train on the combined training and validation set and report results on the test-dev set. Since recall for proposals is not provided by the evaluation server, we train on 118,000 images and report recall on the remaining 5,000 images (commonly referred to as the \textit{minival} set). 

On COCO, we train SNIPER with a batch-size of $128$ and with a learning rate of $0.015$. We use a chip size of $512\times512$ pixels. Training scales are set to ($512$/ms, $1.667$, $3$) where $ms$ is the maximum value width and height of the image\footnote{For the first scale, zero-padding is used if the smaller side of the image becomes less than $512$ pixels.}. The desired area ranges (\textit{i.e.} $\mathcal{R}^i$) are set to ($0$,$80^2$), ($32^2$, $150^2$), and ($120^2$, $\inf$) for each of the scales respectively. Training is performed for a total of $6$ epochs with step-down at the end of epoch $5$. Image flipping is used as a data-augmentation technique. Every epoch requires 11,000 iterations. For training RPN without negatives, each epoch requires 7000 iterations. We use RPN for generating negative chips and train it for $2$ epochs with a fixed learning rate of $0.015$ without any step-down. Therefore, training RPN for $2$ epochs requires less than $20$\% of the total training time. RPN proposals are extracted from all scales. Note that inference takes $1/3$ the time for a full forward-backward pass and we do not perform any flipping for extracting proposals. Hence, this process is also efficient. We use mixed precision training as described in \cite{narang2017mixed}. To this end, we re-scale weight-decay by $100$, drop the learning rate by $100$ and rescale gradients by $100$. This ensures that we can train with activations of half precision (and hence $\sim 2$x larger batch size) without any loss in accuracy. We use fp32 weights for the first convolution layer, last convolution layer in RPN (classification and regression) and the fully connected layers in Faster-RCNN. 

We evaluate SNIPER using a popular detector, Faster-RCNN with ResNets \cite{he2016deep,he2016identity} and MobileNetV2 \cite{mobilenetv2}. Proposals are generated using RPN on top of conv4 features and classification is performed after concatenating conv4 and conv5 features. In the conv5 branch, we use deformable convolutions and a stride of $1$. We use a $512$ dimensional feature map in RPN. For the classification branch, we first project the concatenated feature map to $256$ dimensions and then add $2$ fully connected layers with $1024$ hidden units. For lightweight networks like MobileNetv2 \cite{mobilenetv2}, to preserve the computational processing power of the network, we did not make any architectural changes to the network like changing the stride of the network or added deformable convolutions. We reduced the RPN dimension to $256$ and size of fc layers to $512$ from $1024$. RPN and classification branch are both applied on the layer with stride $32$ for MobileNetv2. 

SNIPER generates $1.2$ million chips for the COCO dataset after the images are flipped. This results in around $5$ chips per image. In some images which contain many object instances, SNIPER can generate as many as $10$ chips and others where there is a single large salient object, it would only generate a single chip. In a sense, it reduces the imbalance in gradients propagated to an instance level which is present in detectors which are trained on full resolution images. At least in theory, training on full resolution images is biased towards large object instances.

\subsection{Recall Analysis}
We observe that recall (averaged over multiple overlap thresholds 0.5:0.05:0.95) for RPN does not decrease if we do not perform negative sampling. This is because recall does not account for false positives. As shown in Section \ref{sec:neg_mining}, this is in contrast to mAP for detection in which negative sampling plays an important role. Moreover, in positive chip sampling, we do cover every ground truth sample. Therefore, for generating proposals, it is sufficient to train on just positive samples. This result further bolsters SNIPER's strategy of finding negatives based on an RPN  in which the training is performed just on positive samples. 

\begin{table}[t]
\begin{center}
\small
\setlength\tabcolsep{5pt}
\begin{tabular}{|c|c|c|c|c|c|c|c|c|c|c|c|}
  \hline
  Method & AR & AR$^{50}$ & AR$^{75}$ & 0-25 & 25-50 & 50-100  & 100-200 & 200-300\\
  \hline\hline
  ResNet-101 With Neg & 65.4 & 93.2 & 76.9 & 41.3 & 65.8 & 74.5 & 76.9 & 78.7 \\
  ResNet-101 W/o Neg & 65.4 & 93.2 & 77.6 & 40.8 & 65.7 & 74.7 & 77.4 & 79.3 \\
  \hline
 \end{tabular}
 \newline
 \caption{We plot the recall for SNIPER with and without negatives. Surprisingly, recall is not effected by negative chip sampling}
\label{tab:rpn_bbox}
\end{center}
\end{table}
\subsection{Negative Chip Mining and Scale}
\label{sec:neg_mining}
SNIPER uses negative chip mining to reduce the false positive rate while speeding up the training by skipping the \textit{easy} regions inside the image. As proposed in Section \ref{sec:neg}, we use a region proposal network trained with a short learning schedule to find such regions. To evaluate the effectiveness of our negative mining approach, we compare SNIPER's mean average precision with a slight variant which only uses positive chips during training (denoted as \textit{SNIPER w/o neg}). All other parameters remain the same. Table \ref{tab:neg_mining} compares the performance of these models. The proposed negative chip mining approach noticeably improves AP for all localization thresholds and object sizes. Noticeably, negative chip mining improves the average precision from $43.4$ to $46.1$. This is in contrast to the last section where we were evaluating proposals. This is because mAP is affected by false positives. If we do not include regions in the image containing negatives which are similar in appearance to positive instances, it would increase our false positive rate and adversely affect detection performance.

SNIPER is an efficient multi-scale training algorithm. In all experiments in this paper we use the aforementioned three scales (See Section \ref{sec:exp_details} for the details). To show that SNIPER effectively benefits from multi-scale training, we reduce the number of scales from $3$ to $2$ by dropping the high resolution scale. Table \ref{tab:neg_mining} shows the mean average precision for SNIPER under these two settings. As can be seen, by reducing the number of scales, the performance consistently drops by a large margin on all evaluation metrics. 

\subsection{Timing}
It takes 14 hours to train SNIPER end to end on a 8 GPU V100 node with a Faster-RCNN detector which has a ResNet-101 backbone. It is worth noting that we train on 3 scales of an image pyramid (max size of 512, 1.667 and 3). Training RPN is much more efficient and it only takes 2 hours for pre-training. Not only is SNIPER efficient in training, it can also process around 5 images per second on a single V100 GPU. For better utilization of resources, we run multiple processes in parallel during inference and compute the average time it takes to process a batch of 100 images.

\begin{table*}[t]
\begin{center}
\small

\begin{tabular}{|c|c|c|c|c|c|c|c|c|c|}
  \hline
  Method & Backbone & AP & AP$^{50}$ & AP$^{75}$ & AP$^{S}$ & AP$^{M}$ & AP$^{L}$ \\
  \hline
  SNIPER   & ResNet-101 & 46.1 & 67.0 & 51.6 & 29.6 & 48.9 & 58.1 \\
  SNIPER 2 scale   & ResNet-101 & 43.3 & 63.7 & 48.6 & 27.1  & 44.7 & 56.1 \\
  SNIPER w/o negatives  & ResNet-101 & 43.4 & 62.8 & 48.8 & 27.4  & 45.2 & 56.2 \\
  \hline   
 \end{tabular}
 \newline
 \caption{The effect training on 2 scales (1.667 and max size of 512). We also show the impact in performance when no negative mining is performed.}
\label{tab:neg_mining}
\end{center}
\end{table*}


\subsection{Inference}
We perform inference on an image pyramid and scale the original image to the following resolutions ($480$, $512$), ($800$, $1280$) and ($1400$, $2000$). The first element is the minimum size with the condition that the maximum size does not exceed the second element. The valid ranges for training and inference are similar to SNIP \cite{singh2017analysis}. For combining the detections, we use Soft-NMS \cite{bodla2017soft}. We do not perform flipping \cite{zagoruyko2016multipath}, iterative bounding box regression \cite{gidaris2016locnet} or mask tightening \cite{liu2018path}.  

\subsection{Comparison with State-of-the-art}
It is difficult to fairly compare different detectors as they differ in backbone architectures (like ResNet \cite{he2016deep}, ResNext \cite{xie2017aggregated}, Xception \cite{chollet2016xception}), pre-training data (\textit{e.g.} ImageNet-5k, JFT \cite{hinton2015distilling}, OpenImages \cite{openimages}), different structures in the underlying network (\textit{e.g} multi-scale features \cite{lin2017feature,najibi2017ssh}, deformable convolutions \cite{dai2017deformable}, heavier heads \cite{peng2017megdet}, anchor sizes, path aggregation \cite{liu2018path}), test time augmentations like flipping, mask tightening, iterative bounding box regression etc. 

Therefore, we compare our results with SNIP \cite{singh2017analysis}, which is a recent method for training object detectors on an image pyramid. The results are presented in Table \ref{tab:final_bbox}. Without using batch normalization \cite{ioffe2015batch}, SNIPER achieves comparable results. While SNIP \cite{singh2017analysis} processes almost all the image pyramid, SNIPER on the other hand, reduces the computational cost by skipping easy regions. Moreover, since SNIPER operates on a lower resolution input, it reduces the memory footprint. This allows us to increase the batch size and unlike SNIP \cite{singh2017analysis}, we can benefit from batch normalization during training. With batch normalization, SNIPER significantly outperforms SNIP in all metrics. It should be noted that not only the proposed method is more accurate, it is also $3\times$ faster during training. To the best of our knowledge, for a Faster-RCNN architecture with a ResNet-101 backbone (with deformable convolutions), our reported result of 46.1\% is state-of-the-art. This result improves to 46.8\% if we pre-train the detector on the OpenImagesV4 dataset. Adding an instance segmentation head and training the detection network along with it improves the performance to 47.6\%.

With our efficient batch inference pipeline, we can process 5 images per second on a single V100 GPU and still obtain an mAP of 47.6\%. This implies that on modern GPUs, it is practical to perform inference on an image pyramid which includes high resolutions like 1400x2000. We also show results for Faster-RCNN trained with MobileNetV2. It obtains an mAP of 34.1\% compared to the SSDLite \cite{mobilenetv2} version which obtained 22.1\%. This again highlights the importance of image pyramids (and SNIPER training) as we can improve the performance of the detector by 12\%. 

\begin{table*}[t]
\begin{center}
\small

\begin{tabular}{|c|c|c|c|c|c|c|c|c|c|}
  \hline
  Method & Backbone & AP & AP$^{50}$ & AP$^{75}$ & AP$^{S}$ & AP$^{M}$ & AP$^{L}$ \\
  \hline
  SSD   & MobileNet-v2 & 22.1 & - & - & - & - & - \\
  \hline
  SNIP   & ResNet-50 (fixed BN) & 43.6 & 65.2 & 48.8 & 26.4 & 46.5 & 55.8 \\
     & ResNet-101 (fixed BN) & 44.4 & 66.2 & 49.9 & 27.3 & 47.4 & 56.9 \\
  \hline
  \hline
  & MobileNet-V2 & 34.1 & 54.4 & 37.7 & 18.2 & 36.9 & 46.2 \\
  & ResNet-50 (fixed BN) & 43.5 & 65.0 & 48.6 & 26.1 & 46.3 & 56.0 \\
SNIPER   & ResNet-101 & 46.1 & 67.0 & 51.6 & 29.6 & 48.9 & 58.1 \\
   & ResNet-101 + OpenImages & 46.8 & 67.4 & 52.5 & 30.5 & 49.4 & 59.6 \\
   & ResNet-101 + OpenImages + Seg Binary & 47.1 & 67.8 & 52.8 & 30.2 & 49.9 & 60.2 \\   
   & ResNet-101 + OpenImages + Seg Softmax & 47.6 & 68.5 & 53.4 & 30.9 & 50.6 & 60.7 \\   
  \hline
  \hline
 SNIPER   & ResNet-101 + OpenImages + Seg Softmax & 38.9 & 62.9 & 41.8 & 19.6 & 41.2 & 55.0 \\ 
 SNIPER   & ResNet-101 + OpenImages + Seg Binary & 41.3 & 65.4 & 44.9 & 21.4 & 43.5 & 58.7 \\   
 \hline
 \end{tabular}
 \newline
 \caption{Ablation analysis and comparison with full resolution training. Last two rows show instance segmentation results when the mask head is trained with N+1 way softmax loss and binary softmax loss for N classes.}
\label{tab:final_bbox}
\end{center}
\end{table*}
We also show results for instance segmentation. The network architecture is same as Mask-RCNN \cite{he2017mask}, just that we do not use FPN \cite{lin2017feature} and use the same detection architecture which was described for object detection. For multi-tasking, we tried two variants of loss functions for training the mask branch. One was a foreground-background softmax function for N classes and another was a N+1 way softmax function. For instance segmentation, the network which is trained with 2-way Softmax loss for each class clearly performs better. But, for object detection, the N+1 way Softmax loss leads to slightly better results. We only use 3 scales during inference and do not perform flipping, mask tightening, iterative bounding-box regression, padding masks before resizing etc. Our instance segmentation results are preliminary and we have only trained 2 models so far.

\section{Related Work}
SNIPER benefits from multiple techniques which were developed over the last year. Notably, it was shown that it is important to train with batch normalization statistics \cite{peng2017megdet,liu2018path,zhao2017pyramid} for tasks like object detection and semantic segmentation. This is one important reason for SNIPER's better performance. SNIPER also benefits from a large batch size which was shown to be effective for object detection \cite{peng2017megdet}. Like SNIP \cite{singh2017analysis}, SNIPER ignores gradients of objects at extreme scales in the image pyramid to improve  multi-scale training.

In the past, many different methods have been proposed to understand the role of context \cite{yu2015multi,bell2016inside,mottaghi2014role}, scale \cite{cai2016unified,yang2016exploit,lin2017feature, najibi2017ssh} and sampling \cite{lin2018focal,shrivastava2016training,boda2017sampling,boda2018universal}. Considerable importance has been given to leveraging features of different layers of the network and designing architectures for explicitly encoding context/multi-scale information \cite{najibi2017ssh,liu2016ssd,zagoruyko2016multipath,zeng2017crafting} for classification. Our results highlight that context may not be very important for training high performance object detectors. 


\section{Conclusion and Future Work}
We presented an algorithm for efficient multi-scale training which sampled low resolution chips from a multi-scale image pyramid to accelerate multi-scale training by a factor of 3 times. In doing so, SNIPER did not compromise on the performance of the detector due to effective sampling techniques for positive and negative chips. As SNIPER operates on re-sampled low resolution chips, it can be trained with a large batch size on a single GPU which brings it closer to the protocol for training image classification. This is in contrast with the common practice of training on high resolution images for instance-level recognition tasks. In future, we would like to accelerate multi-scale inference, because a significant portion of the background can be eliminated without performing expensive computation. It would also be interesting to evaluate at what chip resolution does context start to hurt the performance of object detectors.

\section{Acknowledgement}
The authors would like to thank an Amazon Machine Learning gift for the AWS credits used for this research. The research is based upon work supported by the Office of the Director of National Intelligence (ODNI), Intelligence Advanced Research Projects Activity (IARPA), via DOI/IBC Contract Numbers D17PC00287 and D17PC00345. The U.S. Government is authorized to reproduce and distribute reprints for Governmental purposes not withstanding any copyright annotation thereon. Disclaimer: The views and conclusions contained herein are those of the authors and should not be interpreted as necessarily representing the official policies or endorsements, either expressed or implied of IARPA, DOI/IBC or the U.S. Government.

\medskip
\small
\bibliographystyle{ieee}\bibliography{main.bib}
\end{document}